\documentclass[a4paper,fleqn]{cas-sc}

\usepackage{array}
\usepackage{longtable}
\usepackage{booktabs}
\usepackage{cite}
\usepackage{nomencl}
\usepackage{mdframed}
\usepackage{multicol}
\usepackage{multirow}
\usepackage{epsfig}
\usepackage{amsmath}
\usepackage{amsmath,amssymb,amsfonts}
\usepackage{longtable} % for 'longtable' environment
\usepackage{pdflscape} % for 'landscape' environment
\usepackage{lscape}
\usepackage{amssymb}% http://ctan.org/pkg/amssymb
\usepackage{pifont}% http://ctan.org/pkg/pifont
\usepackage{geometry}
\usepackage{algorithmic}
\usepackage{graphicx}
\usepackage{textcomp}
\usepackage{subfigure}
\usepackage{pstricks}
\usepackage{float}
\usepackage{siunitx}
\usepackage[font=normalsize,labelfont=bf]{caption}
\usepackage{tabularx}
\usepackage[switch]{lineno}
\usepackage[numbers, sort&compress]{natbib}
\usepackage[utf8]{inputenc}
\usepackage{cite}
\usepackage{amsmath,amssymb,amsfonts}
\usepackage{algorithmic}
\usepackage{graphicx}
\usepackage{textcomp}
\usepackage{booktabs}
\usepackage{xcolor,colortbl}
\usepackage{amsmath}
\usepackage{lipsum}
\usepackage{mathtools}
\usepackage{cuted}
\usepackage{ragged2e}
\usepackage{graphicx}
\usepackage{epsfig}
\usepackage{subfigure}
\usepackage{amsmath}
\usepackage{here}
\usepackage{amssymb}
\usepackage{array}
\usepackage{pstricks}
\usepackage[T1]{fontenc}
\usepackage{float}
\usepackage{epsfig}
\usepackage{amsfonts}
\usepackage{siunitx}
\usepackage{cite}
\usepackage{here}
\usepackage[font=normalsize,labelfont=bf]{caption}
\usepackage{tabularx}
\usepackage[switch]{lineno}
\usepackage{epstopdf}
\usepackage{fancyhdr}
\usepackage{eso-pic}
\usepackage{multirow}
 \usepackage{url}
 \usepackage{courier}
 \usepackage[utf8]{inputenc}
\usepackage{longtable}
\usepackage{pdfpages}
\usepackage{tabularray}

\usepackage{caption}
\usepackage{subcaption}

\shortauthors{\textit{A. B. Siddik et al.}}
\title [mode = title]{Comparative Performance of Machine Learning Algorithms for Early Genetic Disorder and Subclass Classification}

\begin{document}

\cortext[cor1]{Corresponding author}

\author[1]{Abu Bakar Siddik}[
prefix=,
role=,
orcid=  ]
\ead[]{abubakar1808031@gmail.com}
\credit{}
\author[1]{Faisal R. Badal}[
prefix=,
role=,
orcid=  ]
\ead[]{faisalrahman1312@gmail.com}
\credit{}
\author[1]{Afroza Islam}[
prefix=,
role=,
orcid=  ]
\fnmark[]
\ead[]{afrozaislam.mte18@gmail.com}
\credit{}

\address[1]{Department of Mechatronics Engineering, Rajshahi University of Engineering \& Technology, Rajshahi 6204, Bangladesh}

\begin{abstract}
A great deal of effort has been devoted to discovering a particular genetic disorder, but its classification across a broad spectrum of disorder classes and types remains elusive. Early diagnosis of genetic disorders enables timely interventions and improves outcomes. This study implements machine learning models using basic clinical indicators measurable at birth or infancy to enable diagnosis in preliminary life stages. Supervised learning algorithms were implemented on a dataset of 22083 instances with 42 features like family history, newborn metrics, and basic lab tests. Extensive hyperparameter tuning, feature engineering, and selection were undertaken. Two multi-class classifiers were developed: one for predicting disorder classes (mitochondrial, multifactorial, and single-gene) and one for subtypes (9 disorders). Performance was evaluated using accuracy, precision, recall, and the F1-score. The CatBoost classifier achieved the highest accuracy of 77\% for predicting genetic disorder classes. For subtypes, SVM attained a maximum accuracy of 80\%. The study demonstrates the feasibility of using basic clinical data in machine learning models for early categorization and diagnosis across various genetic disorders. Applying ML with basic clinical indicators can enable timely interventions once validated on larger datasets. It is necessary to conduct further studies to improve model performance on this dataset.

\end{abstract}


\maketitle

\section{Introduction}

\noindent A genetic disorder is an illness that results from a change or mutation in the DNA (Deoxyribonucleic Acid) sequence that hinders a person from developing normally and healthily \cite{article}. It can be caused by a mutation in one or more genes or by a chromosomal aberration \cite{jain2018chromosomal}. Early and accurate identification of genetic disorders remains an ongoing challenge in healthcare. While significant advances have been made in diagnosing specific conditions, the categorization and prediction of disorders across the spectrum of genetic inheritance types have proven elusive. The ability to systematically discern genetic abnormalities in the preliminary stages of life carries profound clinical implications. Timely diagnosis enables prompt intervention and improves prognosis and quality of life for affected individuals \cite{rasmussen2019alzheimer}. Hence, there is an urgent need for sophisticated yet accessible techniques to delineate the broad classes of genetic disorders and pinpoint particular subtypes.\\

\noindent Machine learning refers to the procedure of acquiring knowledge and skills to develop a statistical model that has the ability to make predictions about future events or classify future observations \cite{callahan2017machine}. In recent years, machine learning (ML) has garnered tremendous interest in advancing genetic research \cite{bracher2021machine, parlett2022applications}, owing to its aptitude for discerning multidimensional interactions devoid of assumptions \cite{ghazal2022supervised}. The supervised learning paradigm has proven especially invaluable for performing robust classification from complex inputs. Supervised learning is a specialized area of machine learning that involves the use of an algorithm to learn from input data that has been explicitly labeled with the aim of producing a desired output. During the training phase, the system is fed sets of data that have been labeled to show the system what values of input correspond to what values of output. Predictions can then be made using the trained model.

\begin{figure*}[h]
    \centering
    \includegraphics[width = 17cm, height=7cm]{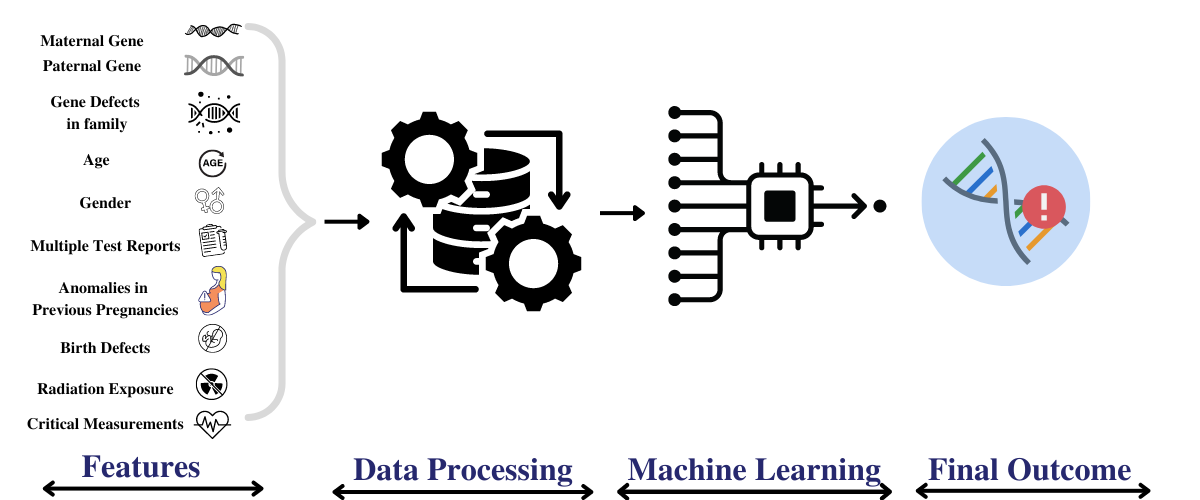}
    \caption{Proposed Workflow}
    \label{fig:workflow}
    \centering
\end{figure*}

\noindent A few pioneering studies have attempted to leverage ML for elucidating the genetic underpinnings of diseases like cancer \cite{iqbal2021clinical}, Diabetes \cite{khanam2021comparison}, Alzheimer's and \cite{kavitha2022early}. However, these works centered on predicting specific illnesses, seldom exploring the full taxonomy. Besides, the predictive models relied predominantly on clinical or imaging data that manifest much later in the disease timeline. To address these limitations, this paper implements an epitomized ML approach for categorizing genetic disorders from baseline indicators observable early in life. We develop two multi-class classifiers using five supervised algorithms - support vector machine (SVM), Random Forest \cite{breiman2001random}, CatBoost \cite{prokhorenkova2017catboost}, Gradient Boosting \cite{friedman2002stochastic}, and LightGBM \cite{ke2017lightgbm}. The K-nearest neighbour (KNN) and Logistic Regression \cite{menard2002applied} algorithms have been excluded from subsequent evaluation due to their suboptimal performance in our initial experimentation. These models have been used to identify the underlying genetic condition (multifactorial, mitochondrial, and single-gene) as well as the specific subtype (Leigh syndrome, Mitochondrial myopathy, Cystic fibrosis, Tay-Sachs, Diabetes, Hemochromatosis, Leber's hereditary optic neuropathy, Alzheimer's disease, and cancer). The input features are derived from readily obtainable parameters like family history, newborn metrics, and basic lab tests. This confers the additional advantage of easy adaptability. Extensive tuning of hyperparameters, feature engineering, and selection are undertaken to optimize model performance.\\

\noindent The key contributions of this work are:
\begin{enumerate}
\item Demonstrating the viability of ML techniques for early delineation across the scope of genetic disorders rather than isolated conditions; 
\item Designing predictive models based solely on elementary clinical variables that can be measured at birth or infancy. 
\end{enumerate}

\noindent The proposed methodology can permit timely interventions and equip families to confront challenges ahead. We envision this pioneering effort to spur more research into data-driven approaches for expediting genetic disorder diagnosis. In the subsequent sections, we describe the dataset, ML architectures, training procedures, evaluation metrics, and results obtained from our experiments.

% \begin{figure*}
% \centering
% \subfigure[]{
%   \includegraphics[width=3in]{files/disorder.png}
%   \label{fig:Genetic_Disorder}
%   }
%  \subfigure[]{
%   \includegraphics[width=3in]{files/subclass.png}
%   \label{fig:Genetic_Subclass}
%   }
% \caption[Optional caption for list of figures]{Class distribution for (a) Genetic Disorder and (b) Genetic Subclass.}
% \label{Genetic Dis and Sub}
% \end{figure*}

\section{Related Works}
\noindent In recent years, machine learning (ML) techniques have shown immense potential for advancing genetic disorder diagnosis and prognosis. ML models are well-suited for discerning complex multivariate relationships from high-dimensional data. For instance, Ghazal et al. \cite{ghazal2022supervised} presents a machine learning approach using SVM and KNN classifiers to predict three diseases - dementia, cancer, and diabetes - from genetic and clinical data.  The SVM model achieved a higher accuracy of 92.8\% on training data and 92.5\% on testing data, compared to KNN which got 92.8\% and 91.2\%. Various statistical measures like sensitivity, specificity, F1-score, etc. were also analyzed. The key contributions are using genetic and clinical data for multiclass disease prediction, testing two standard machine learning models, and achieving state-of-the-art accuracy. In another study Nasir et al. \cite{nasir2022single} proposed a machine-learning approach for the prediction of single gene inheritance disorders (SGID) and mitochondrial gene inheritance disorders (MGID) using patient medical history data. The motivation for both of these papers is that early prediction of these genetic diseases can help improve prognosis and health outcomes and demonstrate the utility of computational intelligence for the early detection of fatal hereditary disorders. Limitations for both of these papers are the lack of model optimization and testing on more genetic markers. \\

\noindent Researchers examined machine learning techniques for predicting psychiatric diseases based on genetic information in \cite{bracher2021machine}. Based on an analysis of multiple studies, the authors have arrived at the conclusion that the effectiveness of diverse machine learning algorithms is subject to variability and that their ultimate performance remains uncertain. Furthermore, it has been found that support vector machines and neural networks are the most commonly utilized machine learning algorithms in such investigations. This study concentrated on the capacity of machine learning techniques to accurately forecast psychiatric disorders solely based on genetic data. Furthermore, they did not emphasize early predictions. \\

\noindent The co-inheritance of DNA variants at two distinct genetic loci has been studied in the context of certain uncommon genetic disorders, such as retinitis pigmentosa and Alport syndrome in \cite{okazaki2022machine}. The authors provide an overview of statistical and machine learning methods for digenic inheritance. Digenic inheritance goes beyond standard Mendelian inheritance where a single genetic variant determines disease status. This study highlights the promise of machine learning to uncover digenic inheritance and gene-gene interactions underlying human disease. However, work is still needed to maximize analytical power while minimizing false discoveries. Mukherjee et al. \cite{mukherjee2021identifying} employed a supervised machine-learning technique and a random forest classifier to identify gene pairings that have the potential to cause digenic disorders. The study compared the functional network and evolutionary features of known digenic gene pairs with real sets of non-digenic gene pairs, including variant pairs from healthy individuals. The aim was to identify gene pairs that could lead to the development of digenic diseases. The findings of the study suggest that the identified gene pairings have the potential to contribute to the development of digenic disorders. \\

\noindent In a study Rahman et al. \cite{rahman2020identification} proposed a machine-learning approach to identify newborns at risk for autism spectrum disorder (ASD) using electronic medical records (EMRs). The authors developed and validated a predictive model based on demographic, clinical, and laboratory features extracted from EMRs of over 200,000 newborns. The model achieved an AUC of 0.81 in the validation cohort and identified several risk factors for ASD, such as male sex, low birth weight, and maternal infections.\\

\noindent Hepatocellular carcinoma (HCC) is the sixth most common cancer in the world. Early diagnosis of HCC is crucial for improving treatment outcomes and reducing the mortality rate. Plawiak et al. \cite{ksikazek2019novel} propose a novel machine learning approach for early detection of HCC patients based on gene expression data. The authors aim to overcome limitations such as high dimensionality, overfitting, noise, and heterogeneity by using a hybrid machine learning approach that combines feature selection, dimensionality reduction, and classification techniques. The authors use gene expression data from 139 HCC patients and 50 healthy controls obtained from the Gene Expression Omnibus (GEO) database. The authors first apply a filter-based feature selection method to select the most relevant genes for HCC prediction. Then, they use a linear discriminant analysis (LDA) method to reduce the dimensionality of the gene expression data and extract the most discriminative features. Finally, they use an SVM method to classify the samples into HCC or non-HCC groups. The authors also use a genetic algorithm to optimize the parameters of the SVM classifier. One of the key limitations of this approach is the use of only gene expression data which may not capture all the biological variations and interactions involved in HCC development and progression.\\

\noindent Iqbal et al. \cite{iqbal2021clinical} provides a review of the clinical applications and future potential of artificial intelligence (AI) and machine learning in cancer diagnosis and treatment. The authors discuss how AI can be used to analyze large datasets to identify patterns and biomarkers to enable early cancer detection, precision diagnosis, and personalized treatment.\\

\noindent While prior studies have made promising advances, some key limitations remain. Most works have focused on predicting specific diseases like cancer, diabetes, and Alzheimer's in isolation rather than categorizing genetic disorders more broadly. The predictive models also tend to rely on clinical, imaging, or molecular data that manifest in later disease stages rather than at birth or infancy. Furthermore, robust validation on large datasets and model optimization is often lacking. Finally, the application of machine learning for expediting early diagnosis across the spectrum of genetic disorders remains relatively unexplored. To address these gaps, this study implements a range of supervised learning models using basic clinical indicators measurable at birth or infancy to categorize genetic disorders at preliminary life stages.

\begin{table}
\centering
\caption{Comparison of Existing Methods with Proposed Method}
\begin{tabular}{|>{\hspace{0pt}}m{0.073\linewidth}|>{\hspace{0pt}}m{0.087\linewidth}|>{\hspace{0pt}}m{0.133\linewidth}|>{\hspace{0pt}}m{0.108\linewidth}|>{\hspace{0pt}}m{0.102\linewidth}|>{\hspace{0pt}}m{0.4\linewidth}|} 
\hline
\textbf{Method}           & \textbf{Dataset}         & \textbf{Features}                                & \textbf{Algorithm}            & \textbf{Performance}      & \textbf{Advantages of Proposed Method}                                                                                                                                                      \\ 
\hline
\textbf{Ghazal et al.}    & Clinical + Genetic Data  & Genetic markers, clinical tests                  & SVM, KNN                      & Accuracy: 92.8\%          & Limited to specific diseases (e.g., cancer, diabetes) and late-stage clinical data.                                                                                                         \\ 
\hline
\textbf{Nasir et al.}     & Patient Medical History  & Genetic inheritance indicators                   & Machine Learning (SGID, MGID) & Early Detection           & Focuses on specific gene disorders, lacks feature diversity.                                                                                                                                \\ 
\hline
\textbf{Mukherjee et al.} & Functional Networks      & Gene pairings for digenic inheritance            & Random Forest                 & Predictive accuracy       & Targets rare digenic disorders, difficult generalization.                                                                                                                                   \\ 
\hline
\textbf{Rahman et al.}    & EMRs                     & Demographics, clinical                           & Logistic Regression, SVM      & AUC: 0.81                 & Focused on autism, limited to EMR-based features.                                                                                                                                           \\ 
\hline
\textbf{Proposed Method}  & Genetic Disorder Dataset & Family history, newborn metrics, basic lab tests & Logistic Regression, CatBoost, Gredientboost, SVM, Random Forest  & Accuracy: 77\% (CatBoost) & Broad coverage of genetic disorder classes, uses early-life measurable features for timely diagnosis. Simplified clinical inputs for early-stage detection and multi-class classification.  \\
\hline
\end{tabular}
\end{table}

\section{Dataset}

\noindent To effectively train and evaluate machine learning models, it is essential to use a dataset that captures meaningful patterns related to genetic disorders. The following subsections provide detailed information on the dataset used in this study, along with the feature engineering and selection techniques applied to optimize model performance.

\subsection{Dataset Description}

\noindent The dataset employed in this study was obtained from Kaggle. The source dataset was a comma-separated file with the majority of columns being categorical and initially consisted of 22083 rows, 42 dependent features, and 2 independent features (genetic disorder and disorder subclass). Table ~\ref{table:features} displays the principal independent features. "Inherited from father" indicates a gene flaw in the patient's father, while "Genes on the mother's side" indicates a gene deficit in the patient's mother. The "Maternal Gene" refers to a genetic defect that originates from the patient's mother, whereas the "Paternal Gene" refers to a genetic fault that originates from the patient's father. The patient's respiration rate is recorded in the "Respiratory Rate (breaths/min)" column, while the heart rate is recorded in the "Heart Rate (rates/min)" column. The "H/O radiation exposure" feature indicates whether or not the patient's parents have a history of radiation exposure, while the "H/O substance abuse" feature indicates whether or not the patient's parents have a history of drug addiction. There are more characteristics such as "History of abnormalities in previous pregnancies" "Number of prior abortions", "Count of White Blood Cells," etc. "Mitochondrial genetic inheritance disorders," "Multifactorial genetic inheritance disorders," and "Single-gene inheritance diseases" are the three categories in the "Genetic Disorder" column, which is one of the two dependent features. "Leigh syndrome," "Mitochondrial myopathy," "Cystic fibrosis," "Tay-Sachs," "Diabetes," "Hemochromatosis," "Leber's hereditary optic neuropathy," "Alzheimer's," and "Cancer" are the nine classes featured in the "Genetic Subclass" column.

\begin{table}[ht]
\centering
\caption{Important Data Features}
\label{table:features}
\begin{tabular}{|c|c|}
\hline
\textbf{Feature Names} & \textbf{Values} \\ \hline
Patient Age & 0-16 \\
Genes in mother's side & 1:yes; 0:no \\
Inherited from father & 1:yes; 0:no \\ 
Maternal gene & 1:yes; 0:no \\ 
Paternal gene & 1:yes; 0:no \\
Blood cell count (mcL) & 4.09-5.609\\ 
Respiratory Rate (breaths/min) & 1:Normal, 0:Tachypnea \\ 
Heart Rate (rates/min) & 1:Normal, 0:Tachycardia \\ 
Birth asphyxia & 1:yes; 0:no \\
Gender & 1:Male; 0:Female; 2:Ambiguous \\
H/O radiation exposure & 1:yes; 0:no; 2:Not applicable \\
H/O substance abuse & 1:yes; 0:no; 2:Not applicable \\
Birth defects & 0:Singular; 1:Multiple \\
\hline
\end{tabular}
\end{table}

\subsection{Data Processing}

\noindent Effective data processing is crucial in preparing the dataset for machine learning models. This process involves cleaning, transforming, and structuring the raw data to enhance the quality and relevance of the features used for classification. By carefully processing the data, we ensure that the models can make accurate and reliable predictions. The following subsections outline the feature engineering and selection techniques applied to improve the model’s performance.

\subsubsection{Feature Engineering}

\noindent Various new features were derived from the original dataset to better capture relevant information. The motivation behind constructing these engineered variables was to amplify pertinent signals related to genetic disorders and handle data sparsity issues. In total, five new engineered features were constructed using techniques like binning, arithmetic operations, and logical rules. All features were motivated by domain insights and intended to better expose predictive signals. The utility of these constructed variables was analyzed in the feature selection stage.\\

\noindent \textbf{Maternal Age Above 40:} A study \cite{shelton2010independent} suggests that a mother's age may increase the risk of autism in her newborn. This binary variable was derived using the thresholding approach.

Let $X_{ma}$ and $X'_{ma}$ be variables for "Maternal Age" and "Maternal Age Above 40".
$X'_{ma} = \begin{cases} 1 & \text{if } age \geq 40 \\ 0 & \text{if } age < 40 \end{cases}$

\noindent \textbf{Number of Symptoms:} This feature was generated by summing the presence of the multiple symptom variables. It aims to quantify the overall symptomatic level of the patient.\\

\noindent \textbf{Any Inherited Gene:} This binary variable checks if any of the two gene inheritance indicators are positive using an OR logical operation. It combines the maternal and paternal inheritance patterns.\\

\noindent \textbf{High WBC Count:} Thresholding was utilized to create this feature to flag abnormally high white blood cell counts based on standard clinical ranges. \\

\noindent \textbf{Heart or Respiratory Issues:} This variable merges two key physiological parameters - heart rate and respiratory rate - using an OR operation to identify any cardiac or respiratory irregularities.

\subsubsection{Feature Selection}
\noindent Feature selection refers to the procedure of selecting a subset of features from an original set of features, guided by specific criteria for feature selection. This identifies the essential characteristics of a dataset. It aids in reducing the amount of data that must be processed by eliminating unnecessary features. Good feature selection results can increase the accuracy of learning, reduce the time required to learn, and make learning results easier to comprehend \cite{cai2018feature}. We incorporated the chi2 feature selection method. It determines the level of similarity of variances between two distributions. The test assumes that the given distributions are independent in its null hypothesis. The mathematical equation for the Chi-Square test is given by:

\begin{equation}
    \chi^2 = \sum_{i=1}^{m} \sum_{j=1}^{k} \frac{(O_{ij}-E_{ij})^2}{E_{ij}}
\end{equation}

\noindent Here, $m$ is the number of attribute values for the feature in question. $k$ is the number of class labels for the output. $O_{ij}$ is the observed frequency and $E_{ij}$ is the expected frequency. For each feature, a contingency table is created with $m$ rows and $k$ columns. Each cell $(i,j)$ denotes the number of rows having attribute feature as $i$ and class label as $k$. Thus each cell in this table denotes the observed frequency. The expected frequency for each cell is calculated by first determining the proportion of the feature value in the total dataset and then multiplying it by the total number of the current class label. The higher the value of $\chi^2$, the more dependent the output label is on the feature and the higher the importance the feature has on determining the output.

\section{Algorithms}

\noindent Various machine learning algorithms have been explored to address the complex nature of genetic disorder classification. Each algorithm brings its unique strengths and challenges when applied to medical datasets. In the following subsections, we discuss the supervised learning algorithms implemented in this study and how they contribute to the classification tasks.

\subsection{Logistic Regression}
\noindent Logistic regression is a commonly employed classification method in the field of machine learning. Logistic regression is characterized by a binary outcome variable, whereas linear regression is characterized by a continuous outcome variable. This is the major distinction between the two types of regression. Because of its tremendous flexibility and understandable interpretation, logistic regression was preferred over alternative distribution functions \cite{hosmer2013applied}. A logistic regression model uses different characteristics to figure out how likely an outcome is \cite{sperandei2014understanding}. The logit function is a fundamental mathematical concept that serves as the basis for logistic regression analysis \cite{peng2002introduction}. The form of the simple logistic model is

\begin{equation}
    logit(Y) = \ln \left(\frac{\pi}{1 - \pi}\right) = \alpha + \beta X
\end{equation}

\noindent The probability of the outcome of interest can be predicted by substituting the antilog of Equation 1 as follows:

\begin{equation}
 \pi(x) = \emph{Probability}(Y = \text{outcome of interest}~ \vert~ X = x, \text{a specific value of} ~X) =  \frac {e^{\alpha + \beta x}}  {1 + e^{\alpha + \beta x}}
\end{equation}

\noindent where \begin{math}\beta \end{math}  represents the regression coefficient, \begin{math}\pi \end{math} represents the probability of the result of interest, and $X$ represents the predictor. This simple logistic regression is extended to multiple predictors (2 predictors) as follows:

\begin{equation}
    logit(Y) = \ln \left(\frac{\pi}{1 - \pi}\right) = \alpha + \beta_{1} X_{1} + \beta_{2} X_{2}
\end{equation}

\noindent Therefore,

\begin{equation}
  \pi(x) = \emph{Probability}(Y = \text{outcome of interest} ~\vert~ X_1 = x_1, X_2 = x_2) =  \frac {e^{\alpha + \beta_{1} x_{1} + \beta_{2} x_{2}}}  {1 + e^{\alpha + \beta_{1} x_{1} + \beta_{2} x_{2}}} 
\end{equation}

\noindent here, the regression coefficients are denoted by \begin{math}\beta_s \end{math}, the probability of the result of interest by \begin{math}\pi \end{math}, the Y intercept by \begin{math} \alpha \end{math} and \begin{math}X_s \end{math} represents the predictors.

\subsection{SVM}
\noindent The Support Vector Machine (SVM) is a widely used machine learning algorithm that can be applied for both classification and regression purposes. The approach is founded upon the concept of Structural Risk Minimization (SRM), thereby endowing it with greater generality. SRM is accomplished by performing an optimization that reduces the maximum value of the generalization error \cite{widodo2007support}. If the training data are linearly separable, we can choose the two margin hyperplanes so that there are no points in between them, and then maximize their distance. Using geometry, we calculate the distance between these two hyperplanes as \begin{math}2/\vert\vert \textbf{w} \vert\vert \end{math}. Given a set of training data \begin{math}D\end{math}, n points of the form
\begin{equation}
    D = \{(\textbf{x}_j, y_j) \: \vert \: \textbf{x}_j \in R^m, y_j \in \{-1, 1\}\}^n_{j = 1}
\end{equation}
where $\textbf{x}_j$ is an \emph{m}-dimensional real vector, \begin{math}y_j\end{math} is either -1 or 1 indicate the class to which point $\textbf{x}_j$ belongs. The minimization of error can be expressed as the quadratic optimization problem that is represented as,

\begin{align*}
    \text{Minimize} :& P(\textbf{w}, b, \xi) = \frac{1}{2} \vert\vert \textbf{w} \vert\vert^2 + C \sum^m_{j=1} \xi_j \\
    \text{Subject to} :& y_j(\{\textbf{w}, \phi (\textbf{x}_j) + b) \geq 1 - \xi_j
\end{align*}

\noindent where $\xi_j \geq 0$ for all integers j between 1 and m, \begin{math}\xi_j \end{math} are slack variables, and C (cost of slack) is a constant. C is a trade-off parameter that determines the optimal margin and training error. The decision function of SVMs can be expressed as \begin{math} f(\textbf{x}) = \textbf{w}^T \phi(\textbf{x}) + bv\end{math}, where the parameters \textbf{w} and b are obtained by solving the optimization problem P as stated in the preceding expression. The optimization problem R can be expressed using Lagrange multipliers as

\begin{align*}
    \text{Minimize}:& F(\beta) = \frac{1}{2}\beta^T\textbf{Q}\beta^T - \beta^T\textbf{1} \\
    \text{Subject to} :&  \textbf{0}\leq \beta \leq \textbf{C}\\ 
    \textbf{y}^T\beta = 0
\end{align*}\

\noindent The notation \begin{math} [Q]_{ij} = y_i y_j\phi^T\: (x_i)\phi(x_j) \end{math} represents the Lagrangian multiplier factor. While familiarity with \begin{math} \phi \end{math} is not mandatory, proficiency in calculating the modified inner product, denoted as the kernel function \begin{math} K(\textbf{x}_i, \textbf{x}_j) = \phi^T(\textbf{x}_i) \phi (\textbf{x}_j)\end{math}, is imperative. As a result, it follows that the expression for \begin{math} [Q]_{ij} \end{math} is given by \begin{math} y_i y_j K(x_i, x_j)\end{math}. According to Mercers' theorem, the optimization problem denoted as P can be classified as a convex quadratic programming (QP) problem that features linear constraints. If the kernel K is positive definite, then the problem can be solved within polynomial time.

\subsection{Random Forest}
\noindent The random forest algorithm is a machine learning technique that comprises a set of predictors. Each tree in the forest predicts its own behavior by considering the values of a random vector that is independently obtained but has the same distribution across all trees in the forest \cite{breiman2001random}. This technique of using a series of predictors to perform one task is known as an ensemble predictor. The ensemble technique used by Random Forest allows it to make more accurate predictions as well as better generalizations \cite{qi2012random}. It is comprised of a collection of tree-structured classifiers denoted by \begin{math}\{h(U,\Theta_k), k= 1, 2, 3, 4,...\} \textnormal{where} \end{math} \begin{math}\{\Theta_k\} \end{math} represents independent identically distributed random vectors. Each tree contributes a single vote towards the most commonly occurring class for the given input \textit{u}. The present study considers a set of classification methods denoted as $h_1(u), h_2(u),..., h_k(u)$, whereby the training set is randomly sampled from the distribution of the random vector \begin{math} V, U \end{math}. The margin function, denoted as \begin{math} mg(U, V) \end{math}, is defined as Equation \ref{eq:mg_uv}.

\begin{equation}\label{eq:mg_uv}
mg(U, V) = av_kI(h_k(U) = V) - max_{j\neq V}av_kI(h_k(U) = j)
\end{equation}

As the number of trees grows, $PE^\ast$ gets closer based on the Equations \ref{eq:p_uv} and \ref{eq:pe_uv}.

\begin{equation}\label{eq:p_uv}
P(U,V)(P\Theta(h(U, \Theta) = V) - max_{j\neq V} P_\Theta(h(U, \Theta) = j)<0
\end{equation}

\begin{equation}\label{eq:pe_uv}
PE^\ast = P(U,V)(mg(U, V)<0)
\end{equation}

\begin{equation}\label{eq:mr_uv}
mr(U,V) = P_\Theta(h(U, \Theta) = V) - max_{j\neq V} P_\Theta(h(U, \Theta) = j)
\end{equation}

\noindent Equation \ref{eq:mr_uv} represents the margin function of the random forest. Equation \ref{eq:s_uv} represents the strength of the set of classifiers $h(U, \theta)$:

\begin{equation}\label{eq:s_uv}
s = E_{U,V}mr(U, V)
\end{equation}

\noindent A more enlightening expression for the variance of $mr$ can be represented as, 

\begin{equation}\label{eq:hat_j_uv}
\hat{j}(U,V) = arg max_{j\neq V} P_\Theta (h(U, \Theta) = j)
\end{equation}

\noindent Thus, Equation \ref{eq:mr_2_uv} illustrates the random forest margin function as:
% \begin{strip}
\begin{equation}\label{eq:mr_2_uv}
mr(U,V) = P_\Theta(h(U, \Theta) = V) - P_\Theta(h(U, \Theta) = \hat{j}(U,V)) = E_\Theta[I(h(U, \Theta) = V) - I(h(U, \Theta) = \hat{j}(U, V))]
\end{equation}
% \end{strip}

\noindent and the raw margin can be expressed as:

\begin{equation}\label{eq:raw_mg_uv}
rmg(\Theta, U, V) = I(h(U, \Theta) = V) - I(h(U, \Theta) = \hat{J}(U, V))
\end{equation}

\noindent The maximum value for the generalization error of the random forest algorithm can be calculated as:

\begin{equation}\label{eq:upper_bound_uv}
PE^\ast \leq \frac{\bar{\rho}(1 - s^2)}{s^2}
\end{equation}

\noindent It has been demonstrated to be highly useful as a method of classification and regression for a variety of applications \cite{biau2016random}. In feature importance measurement, radio frequency (RF) is a widely used methodology. The random forest model provides significant advantages in terms of feature selection owing to its efficient training time, superior accuracy, and lack of requirement for intricate parameter tuning \cite{fei2022cotton}. The random forest has several advantages over typical decision tree methods, including the fact that mature trees are not removed \cite{pal2005random}.

\subsection{Catboost}
\noindent The CatBoost algorithm is an algorithm for machine learning that employs a decision tree model based on the boosting gradient methodology. Gradient boosting refers to the technique of creating a predictor ensemble in multiple dimensions through the use of gradient descent \cite{prokhorenkova2018catboost}. A series of decision trees is built one after the other during training. Each subsequent tree is constructed with less loss than its predecessor. Consider a dataset consisting of instances denoted by \begin{math} \mathcal{Z} = \{(\textbf{x}_t, y_t)\}_{t = 1...n} \textnormal{ where } \textbf{x}_t = x_t^1 ... x_t^m \end{math}, where \begin{math} \textbf{x}_t \end{math} is a random vector of $m$ features and \begin{math} y_t \in \mathbb{R} \end{math} is a target variable that can take on binary or numeric values. The objective of learning tasks is to obtain a function \begin{math} f \colon \mathbb{R}^m \rightarrow \mathbb{R}\end{math} that minimizes the anticipated loss. The function $\mathcal{L}(f)$ is defined as the expected value of the loss function $L(y, f(x))$. The smooth loss function is denoted by $L(..)$, and the test example $(x, y)$ is sampled from $P$, excluding the $\mathcal{Z}$ training set. The gradient boosting technique constructs a series of successive approximations $f^r\colon \mathbb{R}^m \rightarrow \mathbb{R}, r = 0, 1, 2, ...$ in a greedy manner through iterative processes. The current estimate $f^r$ is obtained through an additive derivation process from the previous estimate $f^{r-1}$, as expressed by the equation $ f^r = f^{r-1}+ \alpha h^r $. Here, $\alpha$ denotes the step size, and the function $h^r \colon \mathbb{R}^m \rightarrow \mathbb{R}$, which serves as a base predictor, is selected from a set of functions. The objective is to minimize the expected loss of the variable $H$.

\begin{equation}
    h^r = \mathop{\arg min}_{h \in H} \mathcal{L}(f^{r-1} + h)  = \mathop{\arg min}_{h \in H} \mathbb{E}L(y, f^{r-1}(\textbf{x}) + h(\textbf{x}))
\end{equation}

\noindent The Newton method is commonly employed for the purpose of resolving the minimization problem. This involves utilizing a second-order approximation of $\mathcal{L}(f^{r-1} + h^r)$ at $f^{r-1}$ or a negative gradient step. Both of the mentioned methods are different versions of operational gradient descent. The selection of the gradient step $h^r$ is based on the proximity between $h^r(\textbf{x})$ and $g^r(\textbf{x}, y)$, where $g^r(\textbf{x}, y) \coloneqq \frac{\partial L(y, s)}{\partial s} \vert_{s = f^{r-1}(\textbf{x})}$. A frequent approximation technique is the least-squares method:

\begin{equation}
    h^r = \mathop {\arg min}_{h\in H} \mathbb{E}(-g^r(\textbf{x}, y) - h(\textbf{x}))^2
\end{equation}

\noindent This algorithm is capable of handling categorical features effectively. When choosing the tree structure, it employs a new method for computing leaf values and it reduces overfitting\cite{dorogush2018catboost}.

\subsection{Gradient Boosting}
\noindent Gradient boosting is a machine learning algorithm that has numerous applications, including multiclass classification. It is one of the best ways to build predictive models. 

\noindent The aim of gradient boosting is to derive an estimate, denoted as $\hat{f}(\textbf{x})$, of the function ${f}^\ast (\textbf{x})$ that maps instances $\textbf{x}$ to their corresponding target value $y$, based on a training dataset $\mathcal{Z} = \{\textbf{x}_i, y_i\}^N_1$. This is achieved by minimizing the expected value of a specified loss function, $L(y, f(\textbf{x}))$. The gradient boosting algorithm generates a summation of ${f}^\ast (\textbf{x})$ estimation, which is then multiplied by a weighted combination of functions\\

\begin{equation}
    f_n(\textbf{x}) = f_{n-1}(\textbf{x}) + \rho_n h_n (\textbf{x})
\end{equation}

\noindent where $\rho_n$ is the weight of the $n_{th}$ function, $h_n(\textbf{x})$. These functions represent the ensemble's models (e.g. decision trees). The approximation is constructed in a recursive manner. Initially, a constant estimate of ${f}^\ast (\textbf{x})$ is acquired as

\begin{equation}
    f_0(\textbf{x}) = \mathop {\arg\min}_{\alpha} \sum_{i = 1}^{N} L(y_i, \alpha)
\end{equation}

\noindent Next models are anticipated to minimize
\begin{equation}
    (\rho_n, h_n(\textbf{x})) =\mathop {\arg \min}_{\rho, h} \sum_{i = 1}^{N} L(y_i, f_{n-1}(\textbf{x}_i) + \rho h(\textbf{x}_i ))
\end{equation}

\noindent Instead of directly dealing with the optimization problem, it is possible to view each $h_n$ as a greedy iteration within a gradient descent optimization for $f^\ast$. In this approach, every model $h_n$ undergoes training on a new dataset $\mathcal{Z} = \{\textbf{x}_i, r_{ni}\}^N_{i = 1}$, where the pseudo-residuals, $r_{ni}$, are computed by

\begin{equation}
    r_{ni} = \left[  \frac{ \partial L (y_i, F(\textbf{x})) } {\partial f(\textbf{x})}  \right]_{f(\textbf{x}) = f_{n-1}(\textbf{x})}
\end{equation}

\noindent The determination of the value of $\rho_n$ is achieved through the resolution of an optimization problem involving line search. If the iterative procedure is not sufficiently regularized, there is a risk of overfitting with this approach \cite{bentejac2021comparative}. Gradient boost has the disadvantage of being substantially more time-consuming and inefficient when the data dimension is quite large. This is because they have to look at every piece of data to figure out how much information can be gained from each possible split point.

\subsection{LightGBM}
\noindent Every day, the world is becoming more and more data-driven. 
As the dimension of data grows larger, the gradient boost techniques become more time-consuming. LightGBM was formulated to overcome this issue. LightGBM is a decision tree algorithm that integrates Gradient-based One-Side Sampling (GOSS) and Exclusive Feature Bundling (EFB) with Gradient Boosting Decision Tree (GBDT). Given the supervised training set $X = \{(x_i, y_i)\}^n_{i = 1}$ consists of $n$ samples, where each sample $x$ is associated with a class label $y$. The estimated function is denoted by $F(x)$, and the aim of GBDT optimization is to minimize the loss function $L(y, F(x))$:

\begin{equation}
    \hat{F} = \arg min_F E_{x, y} L(y, F(x))
\end{equation}

\noindent Then, the determination of the iterative criterion of the Gradient Boosting Decision Tree (GBDT) can be achieved through a line search approach aimed at minimizing the loss function as, 

\begin{equation}
    F_m(x) = F_{m-1}(x) + \gamma_m h_m(x)
\end{equation}
where $\gamma_m = \arg \min_\gamma \sum^n_{i = 1} L(y_i, F_{m-1}(x_i) + \gamma h_m(x_i))$, $m$ is the number of iteration, $h_m(x)$ denotes the base decision tree. To separate each node in GBDT, the information gain is commonly used. GOSS is used by LightGBM to calculate variance gain and estimate the split point. Initially, the magnitudes of the gradients pertaining to the training examples are arranged in a descending order. Subsequently, the uppermost $a\times100\%$ data samples, which are denoted as \textbf{A}, are selected based on their gradient values. Subsequently, a stochastic subset denoted as \textbf{B} with cardinality $b\times \lvert A^c \rvert$ is chosen at random from the residual samples $A^c$. The instances are then subdivided based on the estimated variance $V_j^\square(d)$ on $A\cup B$ as, 

\begin{equation}
    V_j^\square(d) = \frac{1}{n}(\frac{(\sum_{x_i \in A_l} g_i + \frac{1-a}{b} \sum_{x_i \in B_l g_i})^2}{n_l^j (d)} + \frac{(\sum_{x_i \in A_r} g_i + \frac{1-a}{b} \sum_{x_i \in B_r g_i})^2}{n_r^j (d)})
\end{equation}

\noindent where $A_l = \{x_i \in A \colon x_{ij} \leq d \}, A_r = \{x_i \in A \colon x_{ij} > d \}, B_l = \{x_i \in B \colon x_{ij} \leq d \}, B_r = \{x_i \in B \colon x_{ij} > d \}, g_i$ denotes the loss function's negative gradient and $\frac{1-a}{b}$ is used to standardize the summation of gradients. The LightGBM algorithm has the potential to expedite the training process by a factor of 20, while maintaining a comparable level of precision \cite{ke2017lightgbm}.

\section{Evaluation}

\noindent Accuracy, precision, recall, f1-score, and confusion matrix have been utilized to evaluate the performance of the model. Before diving into our evaluation system, it's important to understand four different terms.

\begin{enumerate}
    \item \textbf{True positives (TP):} The outcome entails the model's precise prediction of the positive class.
    \item \textbf{True negatives (TN):} The outcome pertains to a scenario where the model has successfully made precise predictions for the negative class.
    \item \textbf{False positives (FP):} It is an outcome where a condition exists when it actually doesn't. It is also known as type-I error.
    \item \textbf{False negatives (FN):} It is a scenario where the model predicts that something is false when in reality it is true. It is the most catastrophic sort of error, commonly known as a type-II error.
\end{enumerate}

\noindent Equations \ref{eq:acc}, \ref{eq:pre}, \ref{eq:re}, and \ref{eq:f1} present the mathematical expressions for Accuracy, Precision, Recall, and F-1 score, respectively.

\begin{equation}\label{eq:acc}
    Accuracy = \frac{TP + TN}{TP + TN + FP + FN}
\end{equation}

\begin{equation}\label{eq:pre}
    Precision = \frac{TP}{TP + FP}
\end{equation}

\begin{equation}\label{eq:re}
    Recall = \frac{TP}{TP + FN}
\end{equation}

\begin{equation}\label{eq:f1}
    F1-score = \frac{2 \times Precision \times Recall}{Precision + Recall}
\end{equation}

% \newcolumntype{L}[1]{>{\centering\let\newline\\\arraybackslash\hspace{0pt}}m{#1}}
% \newcolumntype{C}[1]{>{\centering\let\newline\\\arraybackslash\hspace{0pt}}m{#1}}
% \newcolumntype{R}[1]{>{\centering\let\newline\\\arraybackslash\hspace{0pt}}m{#1}}
% \newcolumntype{S}[1]{>{\centering\let\newline\\\arraybackslash\hspace{0pt}}m{#1}}

% \begin{table} [h!]
% \centering
% \caption{True vs False Positive/negative}
% \label{true vs false}
% \begin{tabular}{cc|c|c|}
% \cline{3-4}
% & & \multicolumn{2}{ c| }{Actual} \\ [1.2ex] \cline{3-4}
% & & Positive & Negative \\ [1.2ex] \cline{1-4}
% \multicolumn{1}{ |c  }{\multirow{3}{*}{\rotatebox{90}{Predicted}} } &
% \multicolumn{1}{ |c| }{Positive} & True Positive & False Positive \\ [1.5ex] \cline{2-4}
% \multicolumn{1}{ |c  }{}                        &
% \multicolumn{1}{ |c| }{Negative} & False Negative & True Negative     \\ [1.5ex]\cline{1-4}
% \end{tabular}
% \end{table}

\section{Result and Discussion}
\noindent The performance of the machine learning models is evaluated using accuracy, precision, recall, and F1-score. The results reveal significant insights into the capabilities and limitations of each model. Table \ref{table:acc} summarizes the overall accuracy of the five implemented classifiers on the genetic disorder and disorder subclass prediction tasks. For categorizing samples into one of the three genetic disorder classes, the CatBoost model achieved the highest accuracy of 77\% on the test set. Support Vector Machine (SVM) also performed well, attaining an accuracy of 76\%. The remaining models had accuracy scores in the 72-75\% range. In the subclass prediction task, SVM emerged as the top performer with 80\% accuracy, slightly outperforming CatBoost at 79\%. The other classifiers had accuracy between 71-73\% for discriminating the 9 different subclasses. The results indicate SVM and CatBoost are the overall best-performing models across both tasks. \\

\begin{table}[b]
\centering
\caption{Overall Accuracy}
\label{table:acc}
\begin{tabular}{|>{\hspace{0pt}}m{0.277\linewidth}|>{\hspace{0pt}}m{0.258\linewidth}|>{\hspace{0pt}}m{0.279\linewidth}|>{\hspace{0pt}}m{0.115\linewidth}|} 
\hline
\textbf{Algorithm} & \textbf{Genetic Disorder} & \textbf{Disorder Subclass} \\ 
\hline
SVM & 0.76 & \textbf{0.80} \\ 
\hline
CatBoost & \textbf{0.77} & 0.79 \\ 
\hline
GradientBoost & 0.72 & 0.71 \\ 
\hline
LGBM & 0.75 & 0.75\\ 
\hline
RandomForest & 0.74 & 0.73\\
\hline
\end{tabular}
\end{table}

\begin{table}
\centering
\caption{Precision of Genetic Disorder Classes}
\label{table:gd pre}
\begin{tabular}{|>{\hspace{0pt}}m{0.294\linewidth}|>{\hspace{0pt}}m{0.225\linewidth}|>{\hspace{0pt}}m{0.213\linewidth}|>{\hspace{0pt}}m{0.196\linewidth}|} 
\hline
\textbf{Algorithms} & \textbf{Mitochondrial} & \textbf{Multifactorial} & \textbf{Single-gene} \\ 
\hline
SVM & 0.59 & \textbf{1.00} & \textbf{0.94} \\ 
\hline
CatBoost & \textbf{0.73} & 0.85 & 0.72 \\ 
\hline
GradientBoost & 0.70 & 0.83 & 0.63 \\ 
\hline
LGBM & 0.71 & 0.84 & 0.69 \\ 
\hline
RandomForest & 0.69 & 0.81 & 0.70 \\
\hline
\end{tabular}
\end{table}

\begin{table}
\centering
\caption{Recall of Genetic Disorder Classes}
\label{table:gd re}
\begin{tabular}{|>{\hspace{0pt}}m{0.294\linewidth}|>{\hspace{0pt}}m{0.225\linewidth}|>{\hspace{0pt}}m{0.213\linewidth}|>{\hspace{0pt}}m{0.196\linewidth}|} 
\hline
\textbf{Algorithms} & \textbf{Mitochondrial} & \textbf{Multifactorial} & \textbf{Single-gene} \\ 
\hline
SVM & \textbf{0.98} & 0.83 & 0.47 \\ 
\hline
CatBoost & 0.75 & \textbf{0.91} & \textbf{0.64} \\ 
\hline
GradientBoost & 0.72 & 0.85 & 0.59 \\ 
\hline
LGBM & 0.79 & 0.87 & 0.58 \\ 
\hline
RandomForest & 0.81 & 0.88 & 0.52 \\
\hline
\end{tabular}
\end{table}

\begin{table}
\centering
\caption{F1-score of Genetic Disorder Classes}
\label{table:gd f1}
\begin{tabular}{|>{\hspace{0pt}}m{0.294\linewidth}|>{\hspace{0pt}}m{0.225\linewidth}|>{\hspace{0pt}}m{0.213\linewidth}|>{\hspace{0pt}}m{0.196\linewidth}|} 
\hline
\textbf{Algorithms} & \textbf{Mitochondrial} & \textbf{Multifactorial} & \textbf{Single-gene} \\ 
\hline
SVM & 0.73 & \textbf{0.91} & 0.63 \\ 
\hline
CatBoost & 0.74 & 0.88 & \textbf{0.67} \\ 
\hline
GradientBoost & 0.71 & 0.84 & 0.61 \\ 
\hline
LGBM & \textbf{0.75} & 0.85 & 0.63 \\ 
\hline
RandomForest & 0.74 & 0.84 & 0.59 \\
\hline
\end{tabular}
\end{table}

\noindent The per-class metrics of precision, recall and F1-score for the genetic disorder classification task are shown in Tables \ref{table:gd pre}-\ref{table:gd f1}. Examining the precision scores, the multifactorial category achieved the highest values across all models, with SVM attaining a perfect precision of 100\%. This demonstrates that the SVM classifier was highly accurate in predicting samples belonging to the multifactorial disorder class. For the mitochondrial and single gene categories, CatBoost and SVM emerged as the top performers with peak precision of 73\% and 94\% respectively. In terms of recall, SVM and CatBoost also achieved the best scores of 98\% and 91\% for the mitochondrial and multifactorial classes. This indicates their ability to correctly retrieve a higher fraction of samples for these classes compared to other models. The single gene category had relatively lower recall, with CatBoost reaching 64\% and no classifier crossing 70\%. The F1-scores follow similar trends as precision and recall, with SVM and CatBoost showing strength for the mitochondrial and multifactorial disorders while performance for the single gene class lags behind. Overall, SVM demonstrates very high precision but lower recall, while CatBoost exhibits a more balanced profile across the classes.\\

\begin{figure*}
\centering
\subfigure[]{
  \includegraphics[width=3.1in, height = 1.7in]{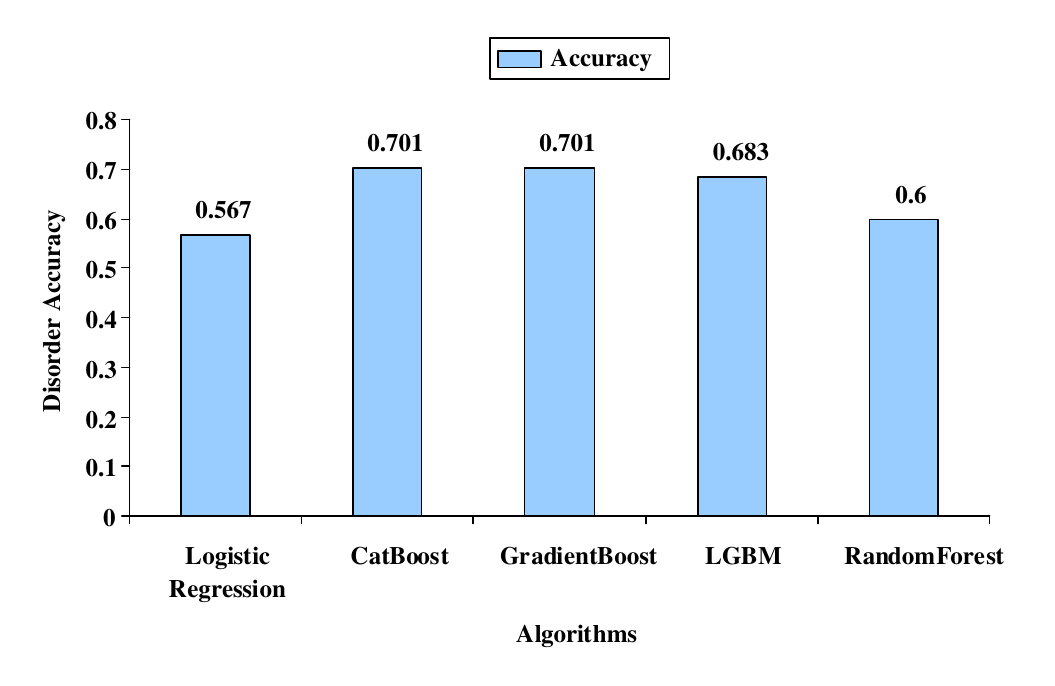}
  \label{fig:acc for gd clf}
  }
 \subfigure[]{
  \includegraphics[width=3.1in, height = 1.7in]{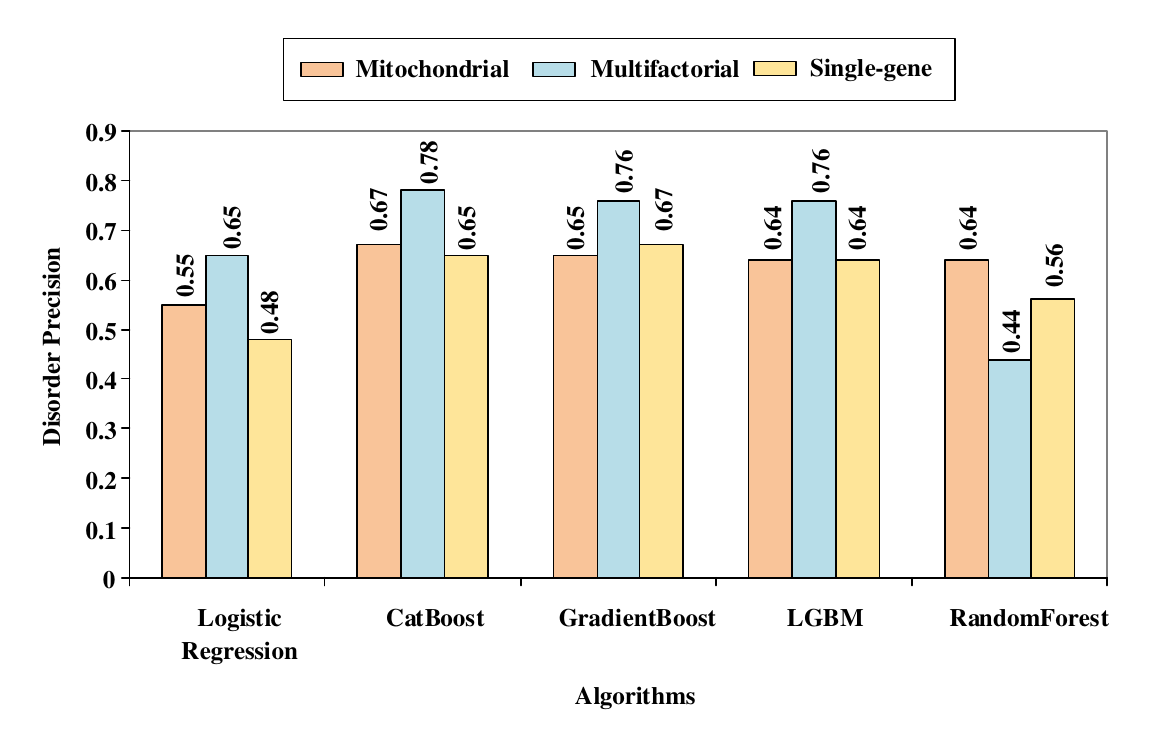}
  \label{fig:precision for gd clf}
  }
\caption[Optional caption for list of figures]{(a) Accuracy and (b) Precision for Genetic Disorder Classifier. }
\label{balanced gd and ds}
\end{figure*}

\begin{figure*}
\centering
\subfigure[]{
  \includegraphics[width=3.1in, height = 1.7in]{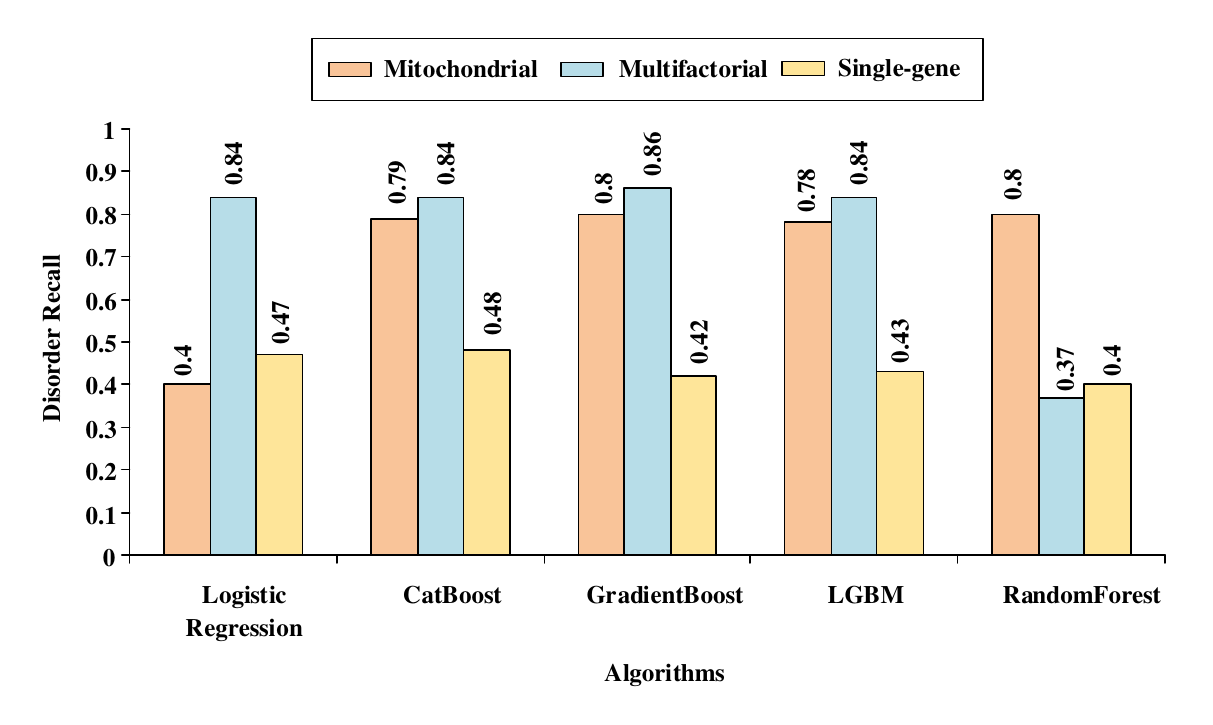}
  \label{fig:recall for gd clf}
  }
 \subfigure[]{
  \includegraphics[width=3.1in, height = 1.7in]{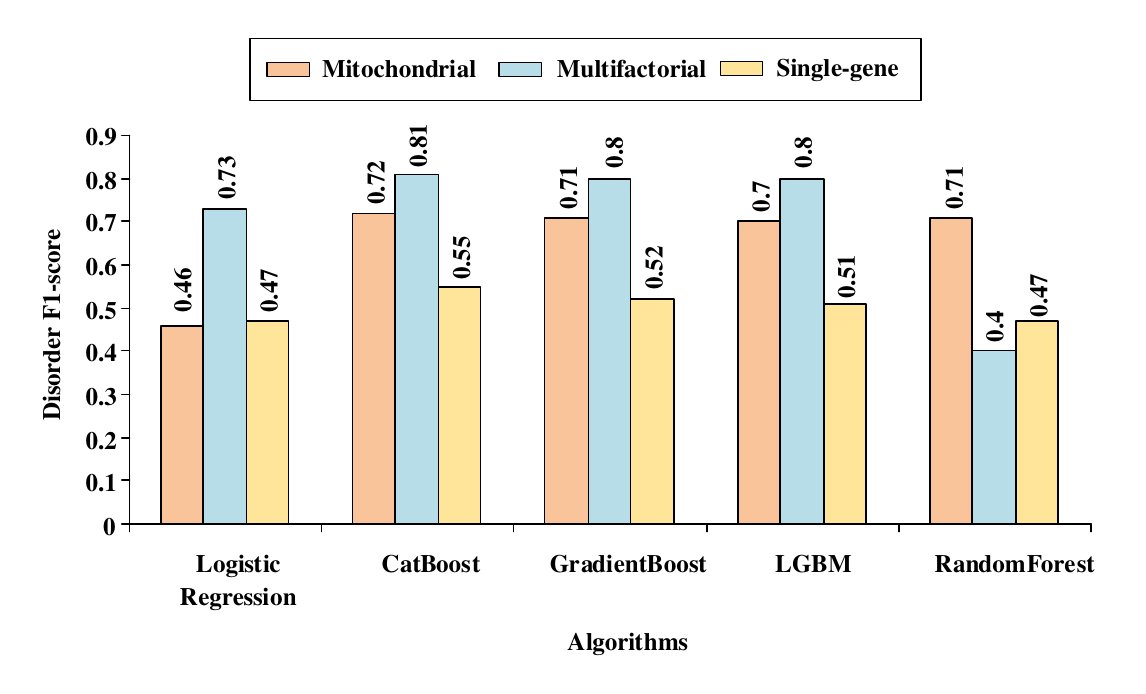}
  \label{fig:f1 for gd clf}
  }
\caption[Optional caption for list of figures]{(a) Recall and (b) F1-score for Genetic Disorder Classifier. }
\label{balanced gd and ds}
\end{figure*}

\begin{figure*}
\centering
\subfigure[]{
  \includegraphics[width=3.1in, height = 1.7in]{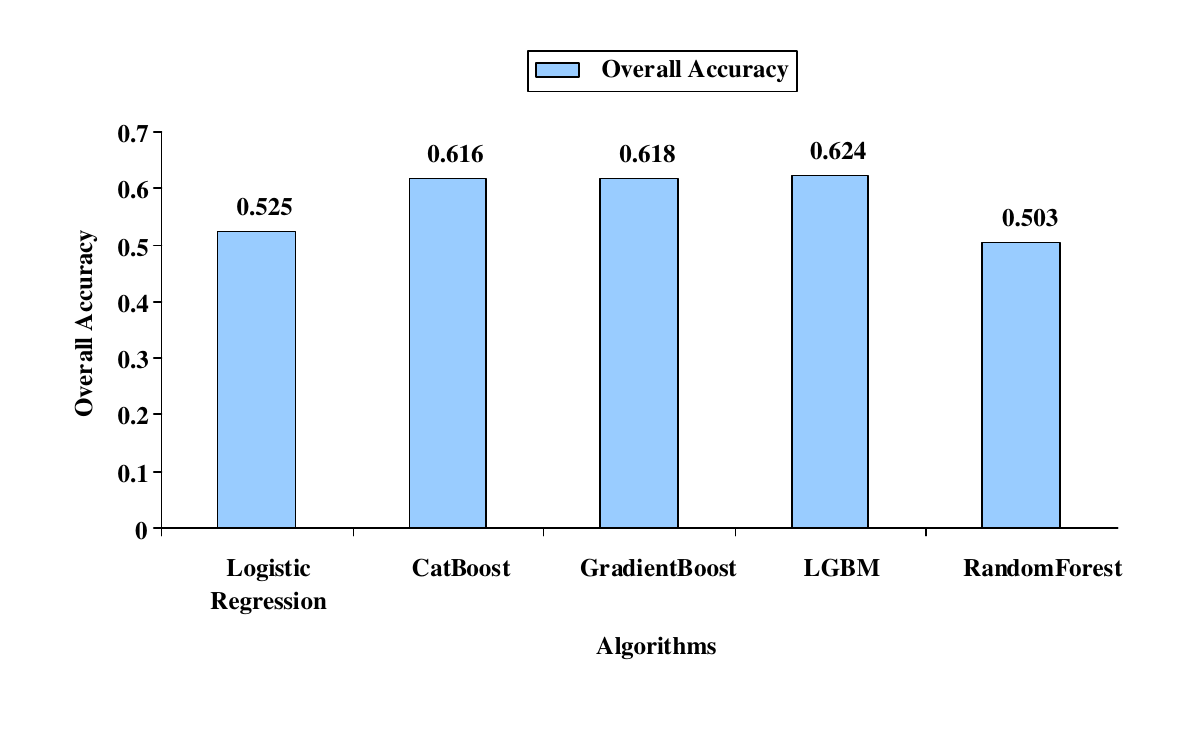}
  \label{fig:overall acc}
  }
 \subfigure[]{
  \includegraphics[width=3.1in, height = 1.7in]{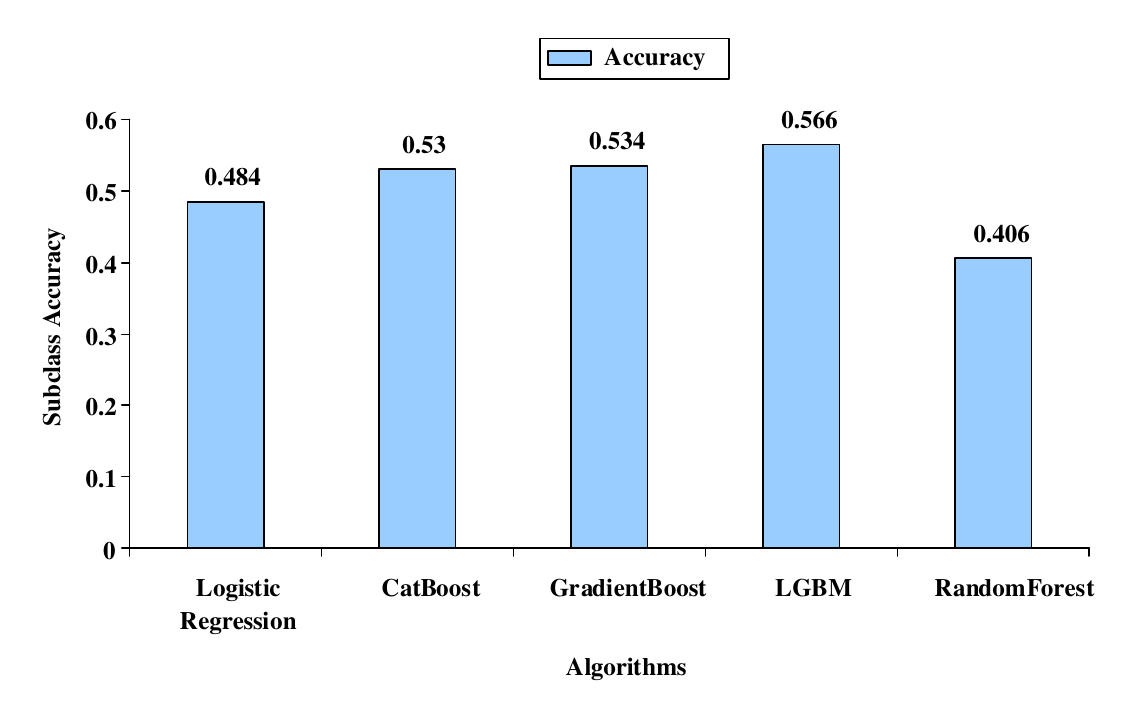}
  \label{fig:acc for ds clf}
  }
\caption[Optional caption for list of figures]{(a) Overall Accuracy and (b) Accuracy for Disorder Subclass Classifier.}
\end{figure*}

\begin{table}
\centering
\caption{Precision of Disorder Subclass}
\label{table:ds pre}
\begin{tabular}{|>{\hspace{0pt}}m{0.20\linewidth}|>{\hspace{0pt}}m{0.17\linewidth}|>{\hspace{0pt}}m{0.106\linewidth}|>{\hspace{0pt}}m{0.16\linewidth}|>{\hspace{0pt}}m{0.075\linewidth}|>{\hspace{0pt}}m{0.13\linewidth}|} 
\hline
\textbf{Subclass} & \textbf{SMV} & \textbf{CatBoost} & \textbf{Gradient Boost} & \textbf{LGBM} & \textbf{Random Forest} \\ 
\hline
Alzheimer's            & 0.99 & 0.99 & 0.98 & \textbf{1.00} & 0.99 \\ 
\hline
Cancer                 & \textbf{1.00} & \textbf{1.00} & 0.99 & 0.99 & 0.99 \\ 
\hline
Cystic fibrosis        & 0.67 & \textbf{0.70} & 0.56 & 0.62 & 0.57 \\ 
\hline
Diabetes               & \textbf{0.86} & 0.85 & 0.72 & 0.77 & 0.77 \\ 
\hline
Hemochromatosis       & \textbf{0.90} & 0.87 & 0.80 & 0.86 & 0.77 \\ 
\hline
LHON                   & \textbf{0.96} & 0.96 & 0.87 & 0.92 & 0.90 \\ 
\hline
Leigh syndrome         & 0.46 & \textbf{0.51} & 0.45 & 0.45 & 0.45 \\ 
\hline
Mitochondrial myopathy & \textbf{0.50} & 0.48 & 0.39 & 0.42 & 0.42 \\ 
\hline
Tay-Sachs              & \textbf{0.78} & 0.73 & 0.64 & 0.71 & 0.69 \\
\hline
\end{tabular}
\end{table}

\begin{table}
\centering
\caption{Recall of Disorder Subclass}
\label{table:ds re}
\begin{tabular}{|>{\hspace{0pt}}m{0.20\linewidth}|>{\hspace{0pt}}m{0.18\linewidth}|>{\hspace{0pt}}m{0.09\linewidth}|>{\hspace{0pt}}m{0.14\linewidth}|>{\hspace{0pt}}m{0.075\linewidth}|>{\hspace{0pt}}m{0.14\linewidth}|} 
\hline
\textbf{Subclass} & \textbf{SVM} & \textbf{CatBoost} & \textbf{Gradient Boost} & \textbf{LGBM} & \textbf{Random Forest} \\ 
\hline
Alzheimer's            & \textbf{1.00} & \textbf{1.00} & 0.99 & 0.99 & 0.99 \\ 
\hline
Cancer                 & \textbf{1.00} & \textbf{1.00} & \textbf{1.00} & \textbf{1.00} & \textbf{1.00} \\ 
\hline
Cystic fibrosis        & \textbf{0.71} & 0.69 & 0.57 & 0.61 & 0.58 \\ 
\hline
Diabetes               & \textbf{0.89} & 0.88 & 0.74 & 0.82 & 0.78 \\ 
\hline
Hemochromatosis       & \textbf{0.94} & 0.90 & 0.81 & 0.84 & 0.85 \\ 
\hline
LHON                   & \textbf{0.99} & 0.96 & 0.86 & 0.90 & 0.90 \\ 
\hline
Leigh syndrome         & 0.41 & \textbf{0.49} & 0.44 & 0.48 & 0.49 \\ 
\hline
Mitochondrial myopathy & \textbf{0.46} & \textbf{0.46} & 0.39 & 0.41 & 0.37 \\ 
\hline
Tay-Sachs              & \textbf{0.78} & 0.74 & 0.63 & 0.68 & 0.62 \\
\hline
\end{tabular}
\end{table}

\begin{table}
\centering
\caption{F1-score of Disorder Subclass}
\label{table:ds f1}
\begin{tabular}{|>{\hspace{0pt}}m{0.20\linewidth}|>{\hspace{0pt}}m{0.18\linewidth}|>{\hspace{0pt}}m{0.09\linewidth}|>{\hspace{0pt}}m{0.14\linewidth}|>{\hspace{0pt}}m{0.075\linewidth}|>{\hspace{0pt}}m{0.14\linewidth}|} 
\hline
\textbf{Subclass} & \textbf{SVM} & \textbf{CatBoost} & \textbf{Gradient Boost} & \textbf{LGBM} & \textbf{Random Forest} \\ 
\hline
Alzheimer's            & \textbf{1.00} & 0.99 & 0.99 & 0.99 & 0.99 \\ 
\hline
Cancer                 & \textbf{1.00} & \textbf{1.00} & 0.99 & 0.99 & 0.99 \\ 
\hline
Cystic fibrosis        & 0.69 & \textbf{0.70} & 0.56 & 0.61 & 0.58 \\ 
\hline
Diabetes               & \textbf{0.87} & 0.86 & 0.73 & 0.80 & 0.77 \\ 
\hline
Hemochromatosis       & \textbf{0.92} & 0.89 & 0.80 & 0.85 & 0.81 \\ 
\hline
LHON                   & \textbf{0.98} & 0.96 & 0.87 & 0.91 & 0.90 \\ 
\hline
Leigh syndrome         & 0.43 & \textbf{0.50} & 0.45 & 0.46 & 0.47 \\ 
\hline
Mitochondrial myopathy & \textbf{0.48} & 0.47 & 0.39 & 0.42 & 0.39 \\ 
\hline
Tay-Sachs              & \textbf{0.78} & 0.74 & 0.63 & 0.69 & 0.65 \\
\hline
\end{tabular}
\end{table}

\noindent The precision, recall, and F1-scores for each of the 9 genetic disorder subclasses are summarized in Tables \ref{table:ds pre}-\ref{table:ds f1}. The metrics showcase wider variability across categories compared to the parent genetic disorder classes. SVM attained perfect precision and recall for Cancer, highlighting robust classification for this subclass. Alzheimer's disease also exhibited high precision and recall exceeding 96\% for all models. On the other end, categories like Leigh Syndrome and Mitochondrial Myopathy proved challenging, with precision and recall struggling to cross 50\% for some classifiers. This performance gap between subclasses emphasizes the difficulty in differentiating rare disorders with subtle phenotypic differences. Examining F1 scores, SVM achieved the top values for most subclasses due to its high precision and recall balance. The inconsistent scores across subclasses and models indicate an opportunity for further tuning focused on hard-to-classify categories. Addressing class imbalance through intelligent oversampling and directing model capacity to minority subclasses could help even outperformance.\\

\noindent Among the implemented models, Support Vector Machine (SVM) and CatBoost emerged as the best performers based on accuracy, precision, recall, and F-1 score metrics as observed earlier. Hence, we further analyzed the Area Under the ROC Curve (AUC) for these two top classifiers. For the genetic disorder prediction task, SVM achieved strong AUC scores of 1.00, 0.86, and 0.86 for the single-gene, mitochondrial, and multifactorial categories respectively. The CatBoost model performed slightly better for the mitochondrial class with an AUC of 0.89, while also attaining 0.98 and 0.83 for the other classes. Examining AUC for the disorder subclasses, both SVM and CatBoost attained perfect scores of 1.00 for Leigh Syndrome, Mitochondrial Myopathy, and Hemochromatosis. This highlights the models' ability to reliably distinguish these rare subclasses. SVM scored marginally higher for LHON (0.80 vs 0.90) and Alzheimer's (0.84 vs 0.89). The classifiers achieved AUC between 0.92-0.99 for Cystic Fibrosis, Tay-Sachs, Diabetes, and Cancer, indicating robust predictive ability. The high AUC values validate the promising classification potential of the developed machine-learning approach across both tasks.

\begin{figure*}[htp]
    \centering
    \subfigure[]{
        \includegraphics[width=3.1in, height=5cm]{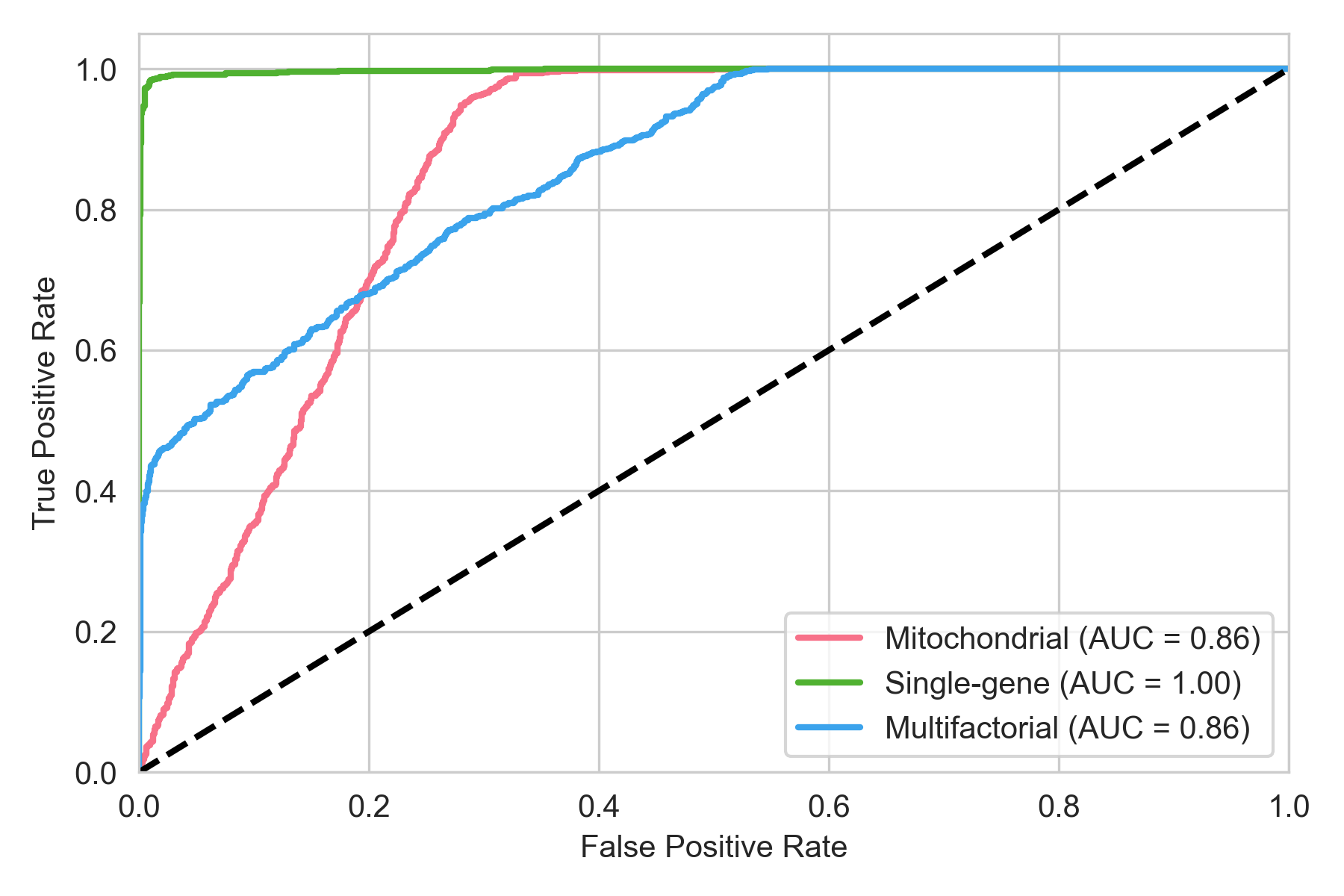}
        % \caption{SVM classifier}
        \label{fig:svm_gd}
        }
    % \hfill
    \subfigure[]{
        \includegraphics[width=3.1in, height=5cm]{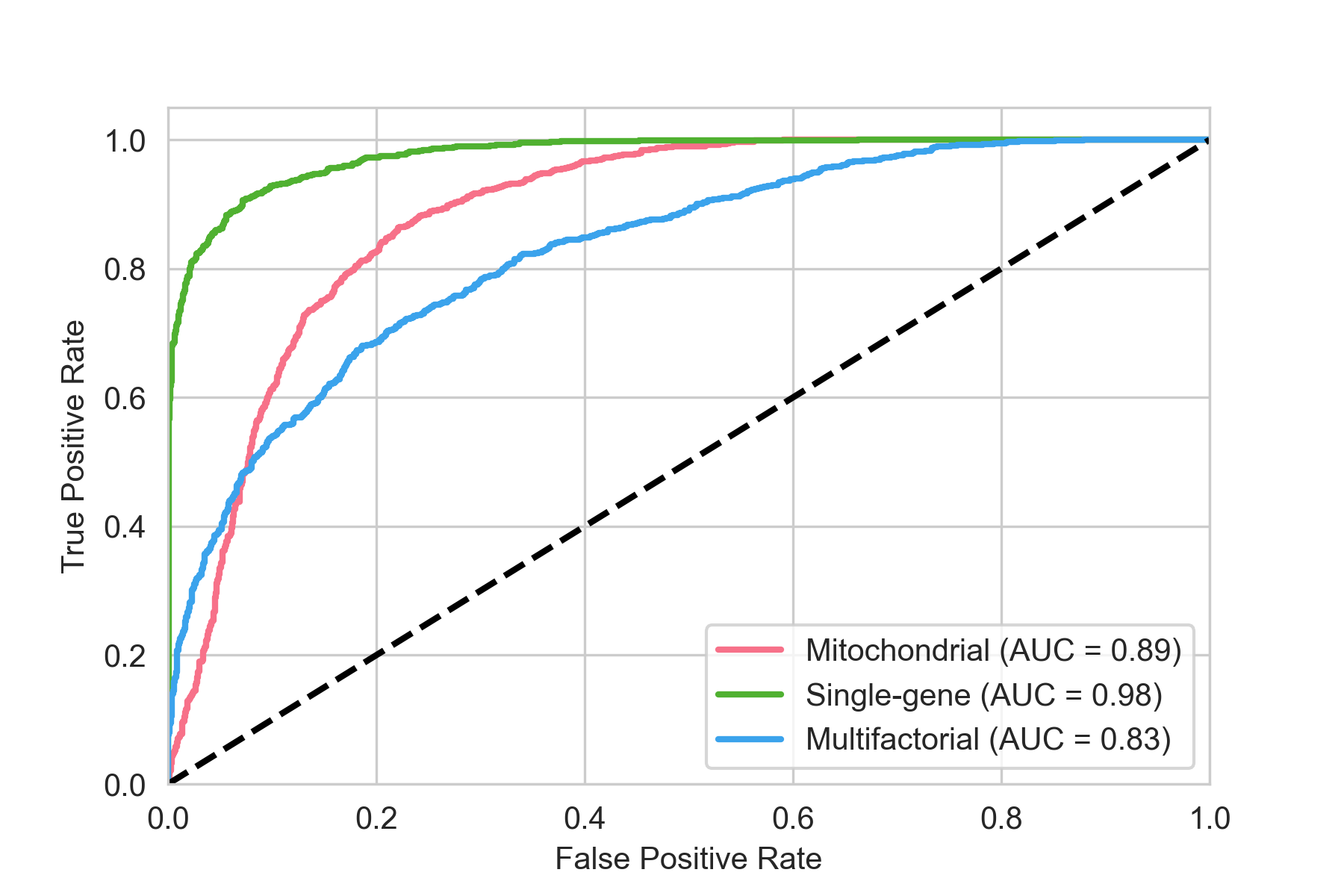}
        % \caption{Catboost classifier}
        \label{fig:cat_gd}
    }
    \caption{ROC curves (a) SVM (b) Catboost classifiers for Genetic Disorder}
    \label{fig:gd_both_images}
\end{figure*}

\begin{figure*}[htp]
    \centering
    \subfigure[]{
        \includegraphics[width=3in, height=5cm]{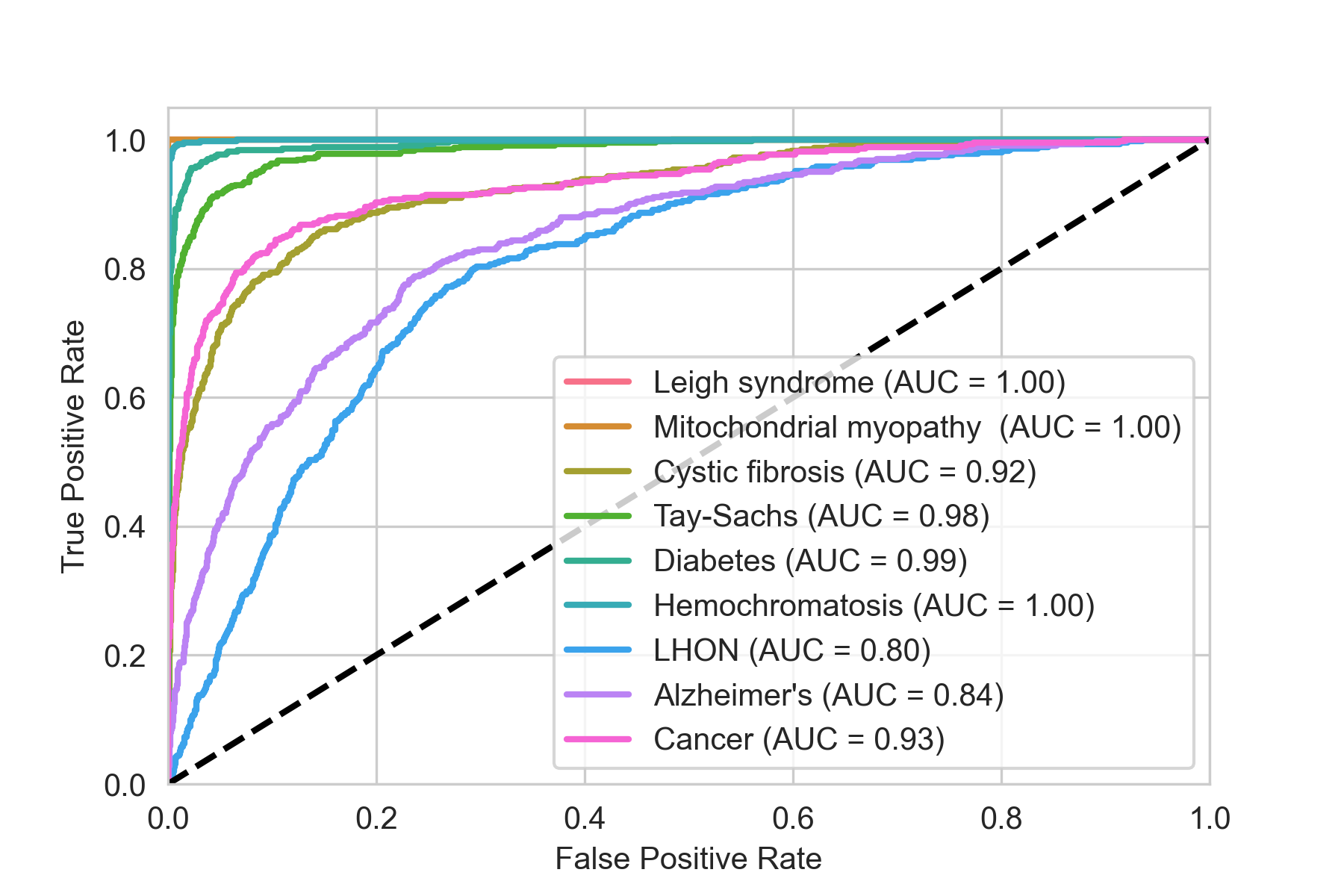}
        % \caption{SVM classifier}
        \label{fig:svm_ds}
    }
    % \hfill
    \subfigure[]{
        \includegraphics[width=3in, height=5cm]{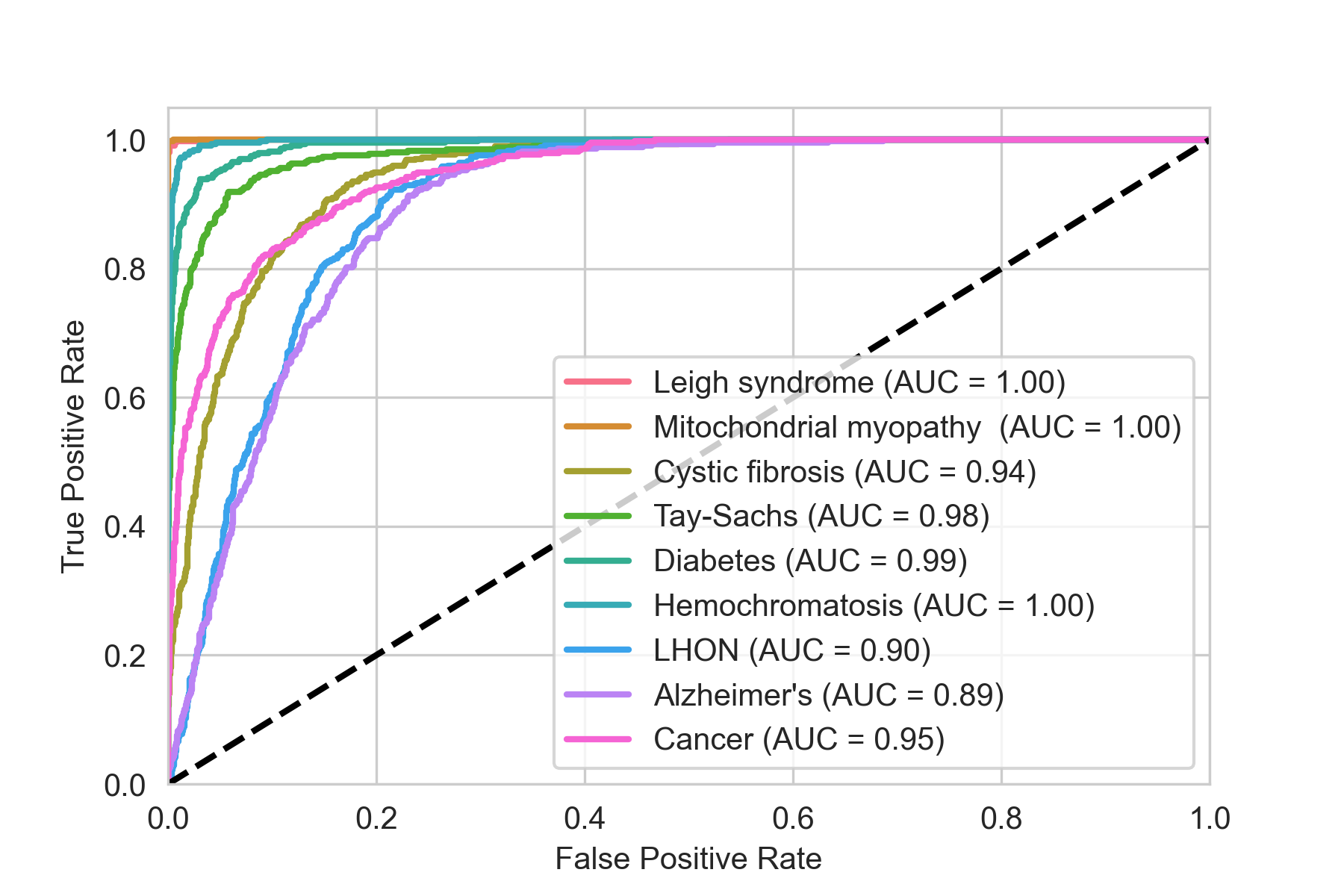}
        % \caption{Catboost classifier}
        \label{fig:cat_ds}
    }
    \caption{ROC curves (a) SVM (b) Catboost classifiers for Disorder Subclass}
    \label{fig:ds_both_images}
\end{figure*}

\section{Conclusion}

\noindent Machine learning is significantly more effective than traditional statistical methods for solving genetic problems. In recent years, it has gained immense popularity in genetics due to its ability to operate in several dimensions and uncover interactions between loci without assuming that they are all identical. This pioneering study demonstrates the potential of machine learning techniques for early classification across the spectrum of genetic disorders using basic clinical indicators. The supervised learning models implemented in this work achieve promising multi-class classification performance, with accuracies reaching 77\% for delineating disorder classes and 80\% for specific subtypes. The results validate the ability to leverage elementary features like family history, newborn metrics, and basic lab tests for expediting diagnosis in preliminary life stages. \\

\noindent By focusing solely on parameters measurable at birth or infancy, the proposed methodology enables timely interventions to improve outcomes. The study provides a framework for augmenting conventional genetic testing with computational intelligence to uncover abnormalities. However, limitations exist regarding model validation across diverse patient populations and real-world clinical integration. \\

\noindent Significant opportunities remain for enhancing model robustness on expanded datasets, addressing class imbalance, and incorporating raw data modalities through advances in deep learning. Overall, this exploratory effort sets the stage for developing machine learning techniques that can make precision medicine more equitable and effective. Further interdisciplinary collaborations between data scientists and geneticists are warranted to translate these approaches into widespread clinical practice. Much work lies ahead, but promising foundations have been laid for intelligent systems to aid in genetic disorder screening and diagnosis.

\section*{Declaration}

\subsection*{Funding}
\noindent Not applicable.

\subsection*{Conflicts of interest/Competing interests}
\noindent Not applicable.

\subsection*{Ethics approval}
\noindent Not applicable.

\subsection*{Consent to participate}
\noindent Not applicable.

\subsection*{Consent for publication}
\noindent Not applicable.

\subsection*{Availability of data and material}
\noindent The data used in this research is publicly available on Kaggle.

\subsection*{Code availability}
\noindent The code used in this research will be made available upon request.

% \subsection*{Authors' contributions}
% The study's design was carried out by ABS, who also developed machine learning models and conducted thorough result analysis. The project was supervised and guided by FRB, who also contributed to the sequence alignment. AF contributed to the sequence alignment and analysis of the results. The manuscript was collaboratively drafted by all the authors. The final manuscript was thoroughly reviewed and approved by all contributing authors.

% \bibliographystyle{unsrt}

\bibliographystyle{elsarticle-num}

\bibliography{main.bib}

\end{document}